\documentclass[
]{ceurart}

\usepackage{listings}
\usepackage{graphicx}
\usepackage{subcaption}
\begin{document}

\copyrightyear{2022}
\copyrightclause{Copyright for this paper by its authors.
  Use permitted under Creative Commons License Attribution 4.0
  International (CC BY 4.0).}

\conference{7th Workshop on Semantic Web Solutions for Large-scale Biomedical Data Analytics}

\title{Integrating Heterogeneous Gene Expression Data through Knowledge Graphs for Improving Diabetes Prediction}


\author[1]{Rita T. Sousa}[%
email=rita.sousa@uni-mannheim.de
]
\cormark[1]
\address[1]{Data and Web Science Group, Universität Mannheim, Germany}
\author[1]{Heiko Paulheim}[%
email=heiko.paulheim@uni-mannheim.de
]

\cortext[1]{Corresponding author.}

\begin{abstract}
  Diabetes is a worldwide health issue affecting millions of people. Machine learning methods have shown promising results in improving diabetes prediction, particularly through the analysis of diverse data types, namely gene expression data. While gene expression data can provide valuable insights, challenges arise from the fact that the sample sizes in expression datasets are usually limited, and the data from different datasets with different gene expressions cannot be easily combined.
  
  This work proposes a novel approach to address these challenges by integrating multiple gene expression datasets and domain-specific knowledge using knowledge graphs, a unique tool for biomedical data integration. KG embedding methods are then employed to generate vector representations, serving as inputs for a classifier. 
  Experiments demonstrated the efficacy of our approach, revealing improvements in diabetes prediction when integrating multiple gene expression datasets and domain-specific knowledge about protein functions and interactions. 
\end{abstract}


\maketitle

\section{Motivation}

Diabetes is a chronic health condition resulting from insufficient insulin production by the pancreas or the body's inability to utilize the insulin it generates effectively~\cite{care2022care}. This disease has emerged as a worldwide health issue, impacting millions of people globally.
According to the World Health Organization, in 2019, diabetes directly contributed to 1.5 million deaths, with 48\% occurring before the age of 70. Besides that, this chronic disease is associated with the development of several comorbidities, such as blindness, kidney failure, heart attacks, strokes, and lower limb amputation. 

Due to the multidisciplinary nature of diabetes, predicting and detecting this complex disease continues to pose a significant challenge.
In the last decades, some approaches have demonstrated encouraging outcomes using machine learning methods to identify patterns and potential risk factors linked to diabetes, allowing not only the early detection of diabetes but also enabling tailored interventions~\cite{jaiswal2021review,sonar2019diabetes,mujumdar2019diabetes,hasan2020diabetes}. 
These machine learning approaches encompass several types of data, including electronic health records~\cite{bertsimas2017personalized}, imaging data~\cite{tang2020prediction}, and demographic data~\cite{xiao2017learning}. Omics data, namely gene expression datasets, have also received attention since genomics, epigenomics, and transcriptomics can help understand the critical pathways and regulatory mechanisms in diabetes~\cite{liu2022uncovering}.

While gene expression datasets are readily accessible in public databases, and gene expression analysis is a powerful tool for pinpointing genes associated with diseases, particularly in the context of diabetes prediction, a significant issue arises in handling this type of data.
On the one hand, gene expression datasets often exhibit a limitation in sample size, with a relatively small number of included samples.
Conversely, supervised machine learning methods are data-driven, relying on a large number of labeled data for effective training and performance.
One alternative involves combining multiple expression datasets to increase the sample pool for training machine learning models. However, this brings us to the challenge of how to integrate the information about multiple expression datasets, as each dataset may measure gene expression across distinct genes. Additionally, variations in experimental platforms and designs across different studies further complicate integration efforts. Knowledge graphs (KGs) present a unique and promising solution. 
KGs can represent knowledge about concepts and relationships in a fully machine-readable format~\cite{hogan2021knowledge}. Moreover, several biomedical ontologies are publicly available to enrich KGs~\cite{rubin2008biomedical}, enabling the representation of domain-specific knowledge. In fact, over the past few years, biomedical ontologies and KGs have emerged as a tool for biomedical data integration and have been adopted in many machine learning applications, with KG embedding approaches~\cite{wang2017knowledge} becoming increasingly popular~\cite{kulmanov2021semantic}.

This work tackles the challenge of integrating heterogeneous gene expression datasets in biomedical applications, focusing on diabetes prediction.
We propose a novel approach that generates a KG to incorporate both gene expression data and domain-specific knowledge and then employs KG embedding methods to generate vector representations of patients.
These patient representations serve as the input for a classifier to predict the likelihood of a patient having diabetes. 
We conducted an evaluation of the impact of integrating multiple gene expression datasets, which showed that incorporating other expression datasets and domain-specific knowledge improves diabetes prediction, emphasizing the efficacy of our approach.
This work is developed in the context of the KI-DiabetesDetektion project, funded by the German Federal Ministry of Education and Research, that aims to integrate biomedical data from various sources and apply machine learning methods to improve the early-stage detection of Diabetes.

\section{Related Work}

Several works have been using gene expression data to predict diabetes, employing diverse methodologies and datasets.
In Li \textit{et al.}~\cite{li2022identification}, a support vector machine classifier is used for the diagnosis of diabetes. While multiple datasets were extracted from the Gene Expression Omnibus database, the machine learning model was trained on only one dataset, with three additional datasets used for validation. Feature selection involved the identification of ten common genes across all datasets.
Mansoori \textit{et al.}~\cite{mansoori2018downregulation} and Kazerouni \textit{et al.}~\cite{kazerouni2020type2} focus on long non-coding RNAs potentially associated with diabetes type 2. Both studies incorporated data collected from 100 diabetic and 100 non-diabetic to train the classifiers.
Mansoori \textit{et al.}~\cite{mansoori2018downregulation} employed logistic regression, whereas Kazerouni \textit{et al.}~\cite{kazerouni2020type2} compare four classifiers ($K$-nearest neighbor, support vector machine, logistic regression, and artificial neural networks) to predict diabetes type 2 using the expression values for specific long non-coding RNAs as input. Both studies suggest that increasing the dataset with a larger number of samples would likely improve the performance of the classifiers.
Furthermore, some other approaches explore expression data for diabetes prediction without employing machine learning methods~\cite{saeidi2018long,zhu2020gene,liu2022uncovering}. 

In the biomedical domain, the exploration of KGs has become increasingly prominent, with KG embedding methods emerging as particularly promising for capturing KG-based information~\cite{chang2020benchmark}. 
These methods map entities and relationships in a KG into a lower-dimensional vector space while preserving graph structure and, in some cases, semantic information. 
Various types of KG embedding methods have been proposed to date. Translational models, exemplified by TransE~\cite{bordes2013translating}, employ distance-based scoring functions to capture relationships between entities. On the other hand, semantic matching approaches, such as distMult~\cite{yang2014embedding}, use similarity-based scoring functions to capture the latent semantics of entities and relations in their vector space representations. Walk-based methods, such as RDF2Vec~\cite{ristoski2016rdf2vec}, employ random walks to generate entity sequences as input to a neural language model that learns latent entity representations. Different walk-based approaches differ in their strategies for random walks and consideration of edge direction and type. In the context of biomedical KGs, characterized by rich hierarchical relations, walk-based approaches emerge as particularly well-suited, considering that these hierarchical relations can be more easily captured in walks.

\section{Methodology}
As discussed above, gene expression datasets typically only have few instances, and different datasets record different gene expressions. Thus, when training prediction models, one can either (1) use only one dataset, thereby having only little training data, or (2) try to combine multiple datasets. In the latter case, those are typically ``incompatible'' in the sense that they have different feature sets, i.e., a naive combination would lead to a larger dataset with lots of NULL values.

\begin{figure}[ht!]
    \centering
    \includegraphics[width=\textwidth]{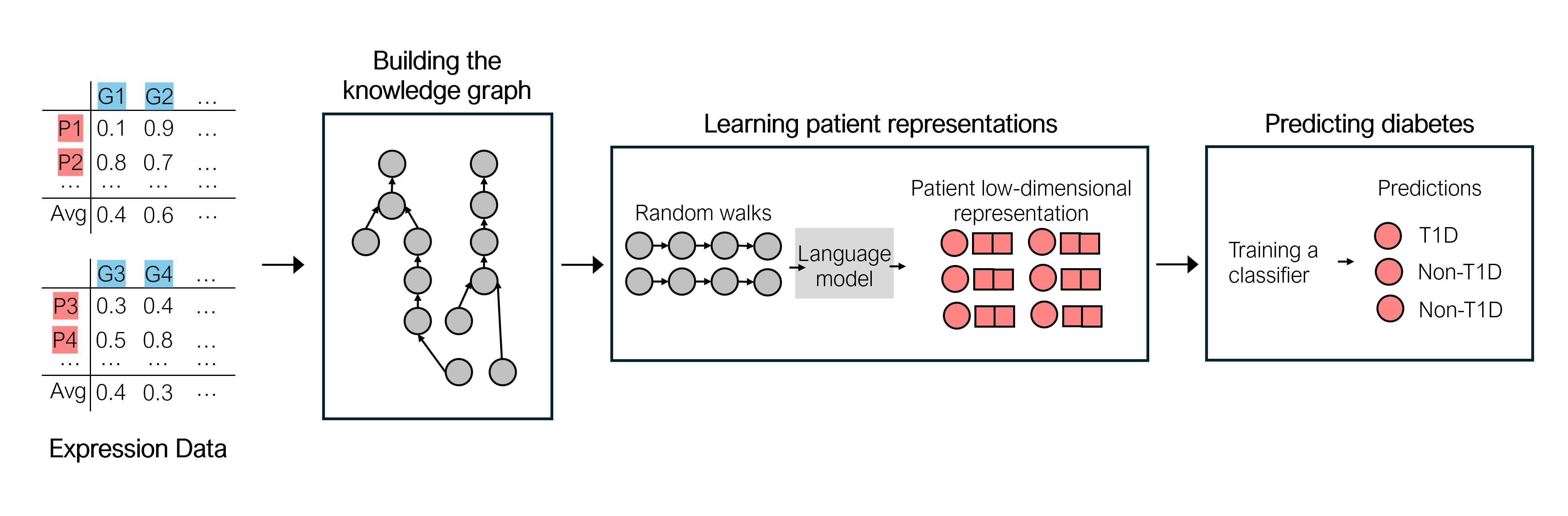}
    \caption{Overview of the proposed methodology with the main steps: building the KG, learning patient representations and predicting diabetes.}
    \label{fig:methodology}
\end{figure}

To overcome these challenges, we propose a methodology to integrate multiple expression datasets into a biomedical KG and then use it for diabetes prediction. Figure~\ref{fig:methodology} shows an overview of this methodology. The first step corresponds to building the KG that integrates not only expression data from different datasets but also domain knowledge on protein function and protein interactions. Then, we generate a vector representation for each patient described in the biomedical KG. The last step involves giving the vectors as input for a classifier. The source code for our methodology is available on GitHub (\url{https://github.com/ritatsousa/expressionKG}).

\subsection{Expression Data}

Several studies have recently explored gene expression for diabetic and non-diabetic individuals, and the findings from these studies can be accessed in publicly available databases.
The Gene Expression Omnibus (GEO)~\cite{clough2024ncbi} is a public database maintained by the National Center for Biotechnology Information that archives high-throughput gene expression and other genomics datasets. 
Each GEO dataset represents a curated collection of biologically comparable GEO samples whose measurements are assumed to be calculated equivalently. The file associated with each dataset contains the raw gene expression data generated by microarrays. In addition to raw data, processed files containing normalized or transformed expression values may be included. In the latter scenario, the data is structured in a tabular format, with each row corresponding to a unique sample, columns representing different genes, and the cells containing specific expression values of those genes for each respective sample.

\subsection{Building the Knowledge Graph}

\begin{figure}
    \centering
    \includegraphics[width=0.9\textwidth]{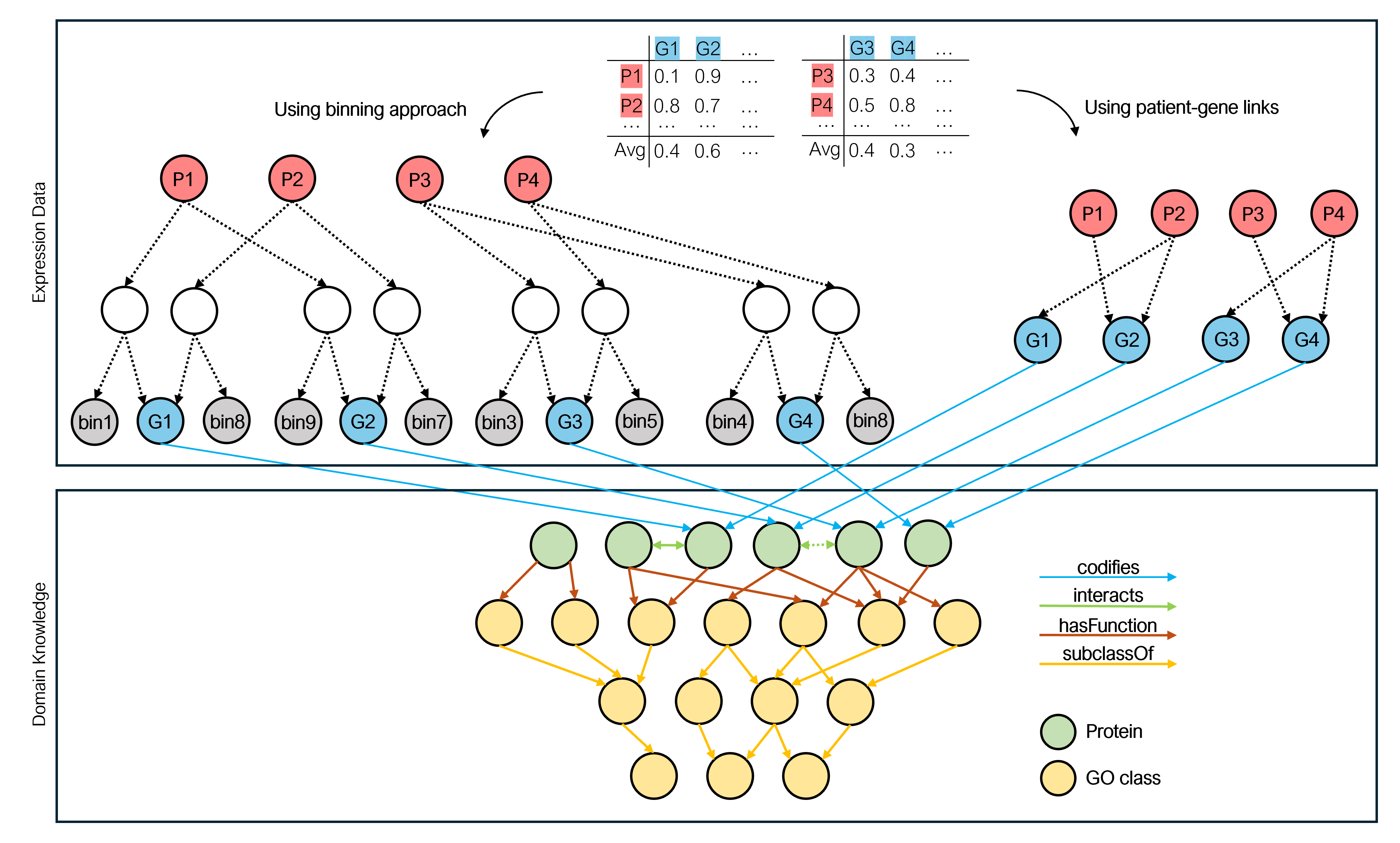}
    \caption{Schema of the two types of data sources and how they are integrated into the KG.}
    \label{fig:datasources}
\end{figure}

The KG is built by integrating two types of data sources: expression data and domain-specific knowledge. Figure~\ref{fig:datasources} illustrates the integration of the two data sources into a KG.

Since our approach relies on KG graph embeddings for generating patient representations and most embedding approaches are not able to handle numeric literals~\cite{preisner2022universal}, we adopt two different strategies to include the expression data in the KG:
\begin{itemize}
    
    \item The first strategy involves representing patient gene expression values in a KG using \textit{blank nodes and binning approaches}. 
    Following the technique proposed in~\cite{preisner2022universal}, we create bins from the set of expression values for each gene within a given dataset. The percentage of unique values defines the number of bins. To implement this, a blank node is generated to represent the expression value attributed to a specific gene for a given patient. This establishes an association wherein a patient is connected to a blank node, which, in turn, is linked to a bin representing the expression value and the corresponding gene.
    Let us consider a simplified example using RDF:
    \par
        \textit{(patientID, rdf:type, :Patient)} \newline 
        \textit{(:geneID,  rdf:type, :Gene)} \newline 
        \textit{(:patientID, :hasExpression, \_:x)} \newline
        \textit{(\_:x, :isExpressionOfGene, :geneID)} \newline
        \textit{(\_:x :hasValue :binID)}
    \par
        \noindent where \textit{\_:x} denotes a blank node.
    
    \item The second strategy employs a \textit{linking approach between patients and genes based on expression values}. A link between a patient and a gene is created when the patient's expression value for that gene is higher than the calculated average expression value for the gene within the dataset. 
\end{itemize}

The domain-specific knowledge includes the Gene Ontology (GO)~\cite{GO2021}, GO annotation data~\cite{GOA2015}, and protein-protein interaction (PPI) data~\cite{string2021}.
The GO defines a hierarchy of classes that describe protein functions that can be represented as a graph where nodes are GO classes and edges define relationships between them.
The GO encompasses three distinct domains for characterizing functions: the biological processes a protein is involved in, the molecular functions a protein performs, and the cellular components where a protein is located. These three domains of GO are represented as separate root ontology classes since they do not share any common ancestor.
The GO annotation data refers to assigning functions represented as GO classes to proteins represented as links in the graph. 
Finally, the PPI data is extracted from STRING~\cite{string2021}, one of the largest available PPI databases that integrates physical interactions and functional associations between proteins collected from several sources. 

To bridge the gap between the two types of data sources, the expression data and the domain-specific knowledge, a gene in the expression data graph is mapped to a protein in the domain-specific KG. Online ID mapping tools, namely UniProt ID Mapping tool\footnote{https://www.uniprot.org/id-mapping}, are used to convert identifiers between genes and proteins. 

\subsection{Learning Patient Representations}

We propose to generate patient representations by leveraging the information of multiple gene expression datasets and domain knowledge. 
As a preliminary step, the KG is converted into a directed and labeled RDF graph, following the W3C's OWL to RDF Graph Mapping guidelines\footnote{https://www.w3.org/TR/owl2-mapping-to-rdf/}. Next, our methodology employs RDF2Vec, a KG embedding method, to generate the low-dimensional vector representations. RDF2Vec~\cite{ristoski2016rdf2vec} is a path-based embedding method that generates random walks in a graph that take into consideration both edge direction and type, making it particularly suited to KGs. Word2vec, a language model, is then employed over random walks on the RDF graph to produce the embeddings. 

Two distinct approaches are employed to represent patients: the first involves generating RDF2vec embeddings directly for the patients using the KG, while the second generates RDF2Vec embeddings for the genes present in gene expression datasets and represents patients as the weighted average of gene embeddings, determined by the respective gene expression values.

\subsection{Predicting Diabetes}

Diabetes prediction is formulated as a binary classification task, where the goal is to categorize a set of patients based on whether they have diabetes or not. Therefore, in the final step, the patient representations are fed into a decision tree~\cite{quinlan1986induction} algorithm for training.

\section{Evaluation}

\subsection{Data}

Three diabetes-related GEO datasets (GSE15932, GSE30208, and GSE55098) are considered for this work (Table~\ref{tab:statisticsdatasets}). These datasets comprise samples associated with two distinct groups: patients diagnosed with type 1 diabetes (T1D) and those serving as control subjects (non-T1D). 
The data from the three datasets are integrated into a KG described in Table~\ref{tab:statisticsKG}.

\begin{table}[h!]
    \caption{Number of samples, number of shared genes across different datasets, and references for each GEO dataset.}
    \begin{tabular}{lllllllllll}
    \toprule
        \multirow{ 2}{*}{\textbf{Dataset}} && \multicolumn{3}{c}{\textbf{Number of samples}}  && \multicolumn{3}{c}{\textbf{Number of shared genes}} && \multirow{ 2}{*}{\textbf{Refs.}} \\  \cmidrule{3-5} \cmidrule{7-9}
        && Total & T1D & non-T1D && GSE30208 & GSE15932 & GSE55098 && \\ \midrule
        GSE30208 &&  63  & 37 & 26 && 368 & 0 & 0 && \cite{GSE30208,kallionpaa2014innate} \\
        GSE15932 &&  22 & 12 & 10 && 0 & 764 & 337 && \cite{GSE15932} \\ 
        GSE55098 &&  16 & 8 & 8 && 0 & 337 & 764 && \cite{GSE55098,yang2015decreased} \\
    \bottomrule
    \end{tabular}
    \label{tab:statisticsdatasets}
\end{table}

\begin{table}[h!]
    \caption{Number of triples, types of relations, GO classes and proteins in the KG.}
    \begin{tabular}{ll}
    \toprule
         & \textbf{Number} \\ \midrule
        Triples & 2433477 \\
        Types of relations & 56 \\ 
        GO classes & 51375 \\
        Proteins & 19169 \\
    \bottomrule
    \end{tabular}
    \label{tab:statisticsKG}
\end{table}

\subsection{Results and Discussion}

\begin{figure}[ht!]
    \centering
    \includegraphics[width=0.65\linewidth]{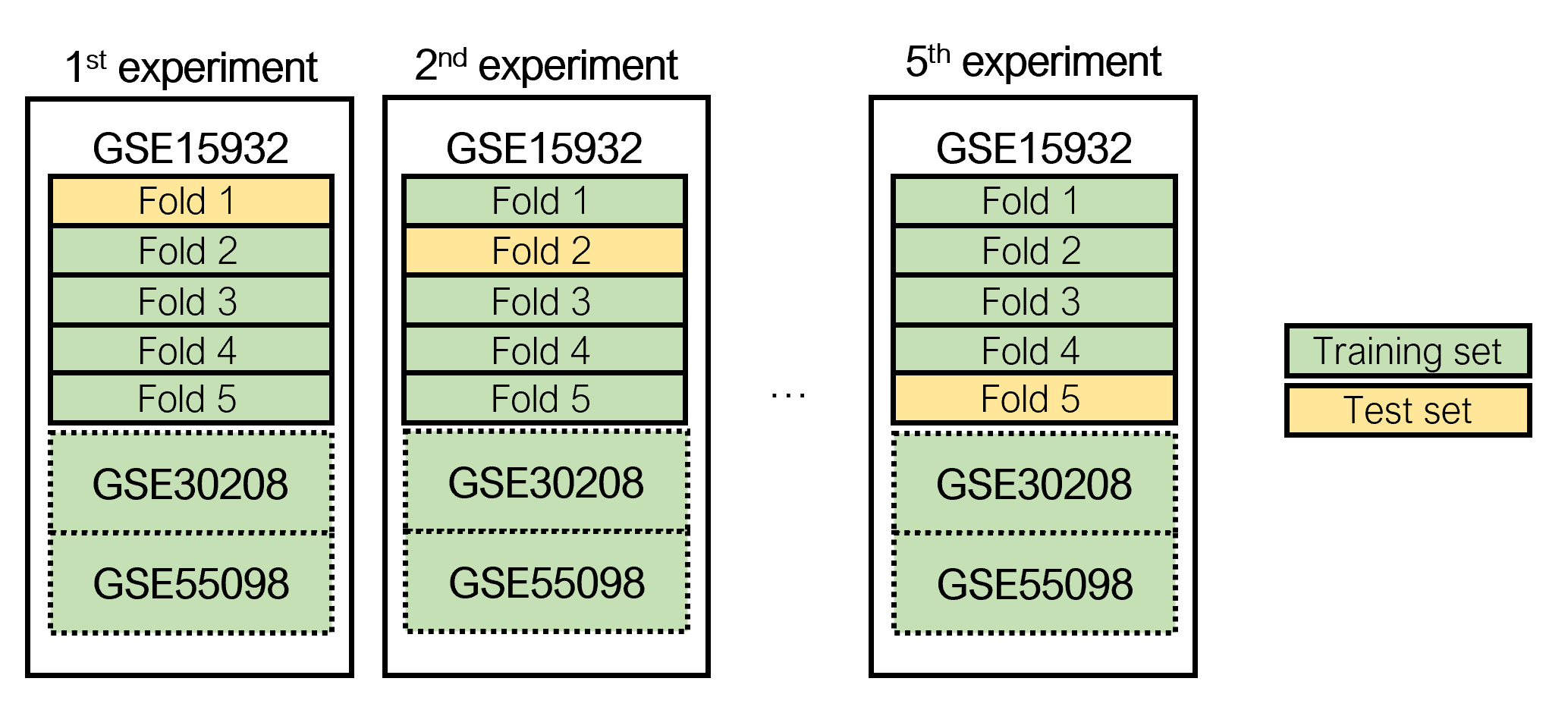}
    \caption{Experimental strategy to split the GSE30208 dataset and enrich with data from the GSE15932 and GSE55098 datasets.}
    \label{fig:experiments}
\end{figure}

To assess the efficacy of the proposed methodology, we analysed the diabetes performance on the GSE30208 dataset by enriching the training data with information from the GSE15932 and GSE55098 datasets.
Since our approach involves integrating data from multiple expression datasets into a KG, we compare it against two baselines that employ the expression values of the patient directly as input for the classifier.
The first baseline exclusively employs data from GSE15932 for training the classifier. The second baseline represents a more simplistic approach to adding information from other datasets. It involves merging all measured genes across datasets and setting the value to 0 when the patient does not have an expression value.
We employed a stratified cross-validation strategy to ensure robust evaluation, dividing the GSE30208 dataset into five folds. The same five folds were used throughout all experiments. The reported results represent the average performance over these five folds. Figure~\ref{fig:experiments} illustrates the employed cross-validation strategy.

Table~\ref{tab:results} shows the accuracy, precision, recall, f-measure, weighted average f-measure and the area under the ROC curve for the baselines and the proposed methodology. 
The two baseline results indicate that simplistically adding information from other datasets does not enhance performance. In fact, it appears to introduce noise to the classifier. This outcome is not unexpected, as the integration of information from diverse datasets is lacking, leading to an ineffective impact on overall performance.
However, by integrating the information from other datasets in a KG, it becomes evident that training a model with diverse datasets improves the performance of machine learning models 
in all metrics, with the exception of precision. Therefore, it confirms our hypothesis that injecting other expression datasets can improve the performance of machine learning models.

However, there are performance variations between the different alternatives of our approach.
For the integration of expression data into the KG, we explore the use of blank nodes and binning approaches versus a linking method based on expression values to link patients and genes. In generating patient representations, we employed two strategies: direct learning of embeddings for patients in the KG; or learning embeddings for genes and representing patients as the weighted average of gene embeddings. This last strategy is independent of the strategy employed to represent the expression data in the KG, so Table~\ref{tab:results} presents only three alternatives. 
Comparing the performance results of Table~\ref{tab:results}, the strategy involving the weighted average of gene embeddings for patient representation emerges as particularly promising because it consistently outperforms the other alternatives. 
Using links between patients and genes based on the expression values is the second-best strategy, and it still improves performance across several metrics compared to the baselines.
Employing the binning approach achieves the worst results, performing worse for many metrics than the baseline. These results may be attributed to the inherent limitations of our path-based embedding method since genes and gene-expression values exist on separate paths.

\begin{table}[t!]
\caption{Average diabetes prediction performance on the GSE30208 dataset for the baselines and the proposed methodology. Acc stands for accuracy, Pr stands for precision, Re stands for recall, F1 stands for f-measure, WAF stands for weighted average f-measure, and AUC stands for area under the ROC curve. For each metric, the best value is in bold.}
\begin{tabular}{llllllllllllll}
\toprule
                                                                          &  \textbf{Acc}  &  \textbf{Pr}  &  \textbf{Re } &  \textbf{F1 } &  \textbf{WAF }   &  \textbf{AUC}   \\ \midrule
\textbf{Baselines} \\
\hspace{0.1cm} Only one dataset   &  0.554     &  \textbf{0.708}      &  0.561   &  0.578     &  0.529  &  0.560  \\
\hspace{0.1cm} Using all the datasets  &  0.442     &  0.650       &  0.314   &  0.396     &  0.422  &  0.474  \\
\textbf{Our methodology} \\
\hspace{0.1cm} Patient rep. using weighted avg. gene emb.  &  \textbf{0.619}     &  0.677      &  \textbf{0.739}   &  \textbf{0.683} &  \textbf{0.589}  &  \textbf{0.606} \\ 
\hspace{0.1cm} Patient rep. using KG with binning approach   &  0.481     &  0.565      &  0.579   &  0.551     &  0.460   &  0.466  \\
\hspace{0.1cm} Patient rep. using KG with patient-gene links &  0.583     &  0.638      &  0.604   &  0.595     &  0.567  &  0.578  \\
\bottomrule
\end{tabular}
\label{tab:results}
\end{table}

\begin{figure}[h!]
    \centering

    \begin{subfigure}{0.495\textwidth}
        \includegraphics[width=\linewidth]{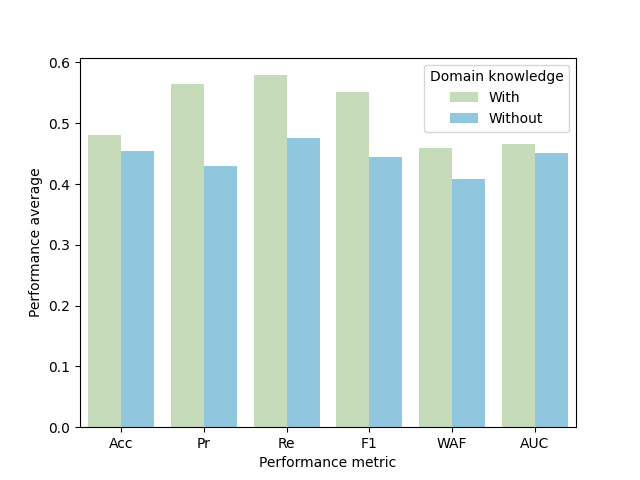}
        \caption{Using binning approach}
        \label{fig:sub1}
    \end{subfigure}
    \hfill
    \begin{subfigure}{0.495\textwidth}
        \includegraphics[width=\linewidth]{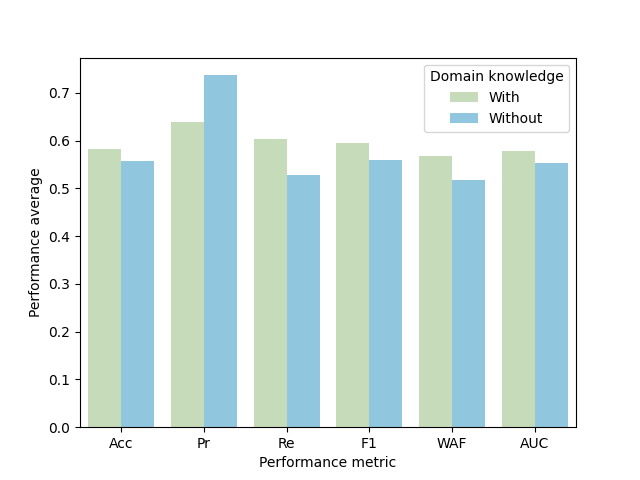}
        \caption{Using patient-gene links}
        \label{fig:sub2}
    \end{subfigure}
    \caption{Performance comparison between using a KG with domain knowledge and without domain knowledge generated with two approaches: binning and patient-gene links. Acc stands for accuracy, Pr stands for precision, Re stands for recall, F1 stands for f-measure, WAF stands for weighted average f-measure, and AUC stands for area under the ROC curve.}
    \label{fig:ablation-studies}
\end{figure}

Since we are interested in investigating the impact of domain-specific knowledge on integrating data from different datasets, we evaluated the diabetes prediction performance using a KG that only contains gene expression data. Figure~\ref{fig:ablation-studies} illustrates the performance variations observed when employing a KG with domain knowledge alongside expression data, compared to utilizing a KG with expression data alone. 
The performance decreases when the domain knowledge is removed for both strategies of building the KG. This demonstrates that knowledge about protein functions and interactions can play an important role in integrating data from datasets measuring gene expression across different genes.

\section{Conclusion}

Several diabetes prediction approaches rely on the analysis of expression data, which provide a detailed molecular profile reflecting gene activity and regulation and therefore can uncover relationships between specific genes and the development of diabetes. However, exploring expression data in machine learning presents its own set of challenges. Existing expression datasets related to diabetes have a very low number of samples what can be a limitation for data-driven methods such as machine learning algorithms. Therefore, the integration of multiple expression datasets can address the issue of limited samples and, at the same time, offer a comprehensive perspective on the complex factors influencing diabetes.

We have developed an approach that enables a comprehensive representation of gene expression data from different datasets within a KG. Through semantic links and domain-specific knowledge, KGs can create a unified knowledge space to connect datasets from distinct studies. In this work, we have explored different strategies to include the expression data in the KG and different strategies to represent the patients within the KG using KG embedding methods. The results of our experiments showed that integrating gene expression data in a KG is able to improve the performance of diabetes prediction. 

The proposed approach is versatile and can be extended to the prediction of other diseases. In addition, since graph neural networks has gained substantial traction recently, as future work, we aim to investigate how can these architectures explicitly designed for graph structures can be used rather than the conventional process of generating embeddings and given them as input for classical machine learning methods such as decision trees.

\begin{acknowledgments}
  The work presented in this paper has been partly funded by the German Federal Ministry of Education and Research under grant number 13GW0661C (KI-DiabetesDetektion).
\end{acknowledgments}

\bibliography{sample-ceur}

\begin{thebibliography}{33}
\expandafter\ifx\csname natexlab\endcsname\relax\def\natexlab#1{#1}\fi
\providecommand{\url}[1]{\texttt{#1}}
\providecommand{\href}[2]{#2}
\providecommand{\path}[1]{#1}
\providecommand{\DOIprefix}{doi:}
\providecommand{\ArXivprefix}{arXiv:}
\providecommand{\URLprefix}{URL: }
\providecommand{\Pubmedprefix}{pmid:}
\providecommand{\doi}[1]{\href{http://dx.doi.org/#1}{\path{#1}}}
\providecommand{\Pubmed}[1]{\href{pmid:#1}{\path{#1}}}
\providecommand{\bibinfo}[2]{#2}
\ifx\xfnm\relax \def\xfnm[#1]{\unskip,\space#1}\fi
\bibitem[{Care(2022)}]{care2022care}
\bibinfo{author}{D.~Care},
\newblock \bibinfo{title}{Care in diabetes—2022},
\newblock \bibinfo{journal}{Diabetes care} \bibinfo{volume}{45} (\bibinfo{year}{2022}) \bibinfo{pages}{S17}.
\bibitem[{Jaiswal et~al.(2021)Jaiswal, Negi, and Pal}]{jaiswal2021review}
\bibinfo{author}{V.~Jaiswal}, \bibinfo{author}{A.~Negi}, \bibinfo{author}{T.~Pal},
\newblock \bibinfo{title}{A review on current advances in machine learning based diabetes prediction},
\newblock \bibinfo{journal}{Primary Care Diabetes} \bibinfo{volume}{15} (\bibinfo{year}{2021}) \bibinfo{pages}{435--443}.
\bibitem[{Sonar and JayaMalini(2019)}]{sonar2019diabetes}
\bibinfo{author}{P.~Sonar}, \bibinfo{author}{K.~JayaMalini},
\newblock \bibinfo{title}{Diabetes prediction using different machine learning approaches},
\newblock in: \bibinfo{booktitle}{2019 3rd International Conference on Computing Methodologies and Communication (ICCMC)}, \bibinfo{organization}{IEEE}, \bibinfo{year}{2019}, pp. \bibinfo{pages}{367--371}.
\bibitem[{Mujumdar and Vaidehi(2019)}]{mujumdar2019diabetes}
\bibinfo{author}{A.~Mujumdar}, \bibinfo{author}{V.~Vaidehi},
\newblock \bibinfo{title}{Diabetes prediction using machine learning algorithms},
\newblock \bibinfo{journal}{Procedia Computer Science} \bibinfo{volume}{165} (\bibinfo{year}{2019}) \bibinfo{pages}{292--299}.
\bibitem[{Hasan et~al.(2020)Hasan, Alam, Das, Hossain, and Hasan}]{hasan2020diabetes}
\bibinfo{author}{M.~K. Hasan}, \bibinfo{author}{M.~A. Alam}, \bibinfo{author}{D.~Das}, \bibinfo{author}{E.~Hossain}, \bibinfo{author}{M.~Hasan},
\newblock \bibinfo{title}{Diabetes prediction using ensembling of different machine learning classifiers},
\newblock \bibinfo{journal}{IEEE Access} \bibinfo{volume}{8} (\bibinfo{year}{2020}) \bibinfo{pages}{76516--76531}.
\bibitem[{Bertsimas et~al.(2017)Bertsimas, Kallus, Weinstein, and Zhuo}]{bertsimas2017personalized}
\bibinfo{author}{D.~Bertsimas}, \bibinfo{author}{N.~Kallus}, \bibinfo{author}{A.~M. Weinstein}, \bibinfo{author}{Y.~D. Zhuo},
\newblock \bibinfo{title}{Personalized diabetes management using electronic medical records},
\newblock \bibinfo{journal}{Diabetes care} \bibinfo{volume}{40} (\bibinfo{year}{2017}) \bibinfo{pages}{210--217}.
\bibitem[{Tang et~al.(2020)Tang, Gao, Lee, Wells, Spann, Terry, Carr, Huo, Bao, and Landman}]{tang2020prediction}
\bibinfo{author}{Y.~Tang}, \bibinfo{author}{R.~Gao}, \bibinfo{author}{H.~H. Lee}, \bibinfo{author}{Q.~S. Wells}, \bibinfo{author}{A.~Spann}, \bibinfo{author}{J.~G. Terry}, \bibinfo{author}{J.~J. Carr}, \bibinfo{author}{Y.~Huo}, \bibinfo{author}{S.~Bao}, \bibinfo{author}{B.~A. Landman},
\newblock \bibinfo{title}{Prediction of type ii diabetes onset with computed tomography and electronic medical records},
\newblock in: \bibinfo{booktitle}{Multimodal Learning for Clinical Decision Support and Clinical Image-Based Procedures: 10th International Workshop, ML-CDS 2020, and 9th International Workshop, CLIP 2020, Held in Conjunction with MICCAI 2020}, \bibinfo{organization}{Springer}, \bibinfo{year}{2020}, pp. \bibinfo{pages}{13--23}.
\bibitem[{Xiao et~al.(2017)Xiao, Gao, Vu, and Turaga}]{xiao2017learning}
\bibinfo{author}{H.~Xiao}, \bibinfo{author}{J.~Gao}, \bibinfo{author}{L.~Vu}, \bibinfo{author}{D.~S. Turaga},
\newblock \bibinfo{title}{Learning temporal state of diabetes patients via combining behavioral and demographic data},
\newblock in: \bibinfo{booktitle}{Proceedings of the 23rd ACM SIGKDD International Conference on Knowledge Discovery and Data Mining}, \bibinfo{year}{2017}, pp. \bibinfo{pages}{2081--2089}.
\bibitem[{Liu et~al.(2022)Liu, Liu, Yu, Qiu, Jiang, and Li}]{liu2022uncovering}
\bibinfo{author}{J.~Liu}, \bibinfo{author}{S.~Liu}, \bibinfo{author}{Z.~Yu}, \bibinfo{author}{X.~Qiu}, \bibinfo{author}{R.~Jiang}, \bibinfo{author}{W.~Li},
\newblock \bibinfo{title}{Uncovering the gene regulatory network of type 2 diabetes through multi-omic data integration},
\newblock \bibinfo{journal}{Journal of Translational Medicine} \bibinfo{volume}{20} (\bibinfo{year}{2022}) \bibinfo{pages}{604}.
\bibitem[{Hogan et~al.(2021)Hogan, Blomqvist, Cochez, d’Amato, Melo, Gutierrez, Kirrane, Gayo, Navigli, Neumaier et~al.}]{hogan2021knowledge}
\bibinfo{author}{A.~Hogan}, \bibinfo{author}{E.~Blomqvist}, \bibinfo{author}{M.~Cochez}, \bibinfo{author}{C.~d’Amato}, \bibinfo{author}{G.~D. Melo}, \bibinfo{author}{C.~Gutierrez}, \bibinfo{author}{S.~Kirrane}, \bibinfo{author}{J.~E.~L. Gayo}, \bibinfo{author}{R.~Navigli}, \bibinfo{author}{S.~Neumaier}, et~al.,
\newblock \bibinfo{title}{Knowledge graphs},
\newblock \bibinfo{journal}{ACM Computing Surveys (Csur)} \bibinfo{volume}{54} (\bibinfo{year}{2021}) \bibinfo{pages}{1--37}.
\bibitem[{Rubin et~al.(2008)Rubin, Shah, and Noy}]{rubin2008biomedical}
\bibinfo{author}{D.~L. Rubin}, \bibinfo{author}{N.~H. Shah}, \bibinfo{author}{N.~F. Noy},
\newblock \bibinfo{title}{Biomedical ontologies: a functional perspective},
\newblock \bibinfo{journal}{Briefings in bioinformatics} \bibinfo{volume}{9} (\bibinfo{year}{2008}) \bibinfo{pages}{75--90}.
\bibitem[{Wang et~al.(2017)Wang, Mao, Wang, and Guo}]{wang2017knowledge}
\bibinfo{author}{Q.~Wang}, \bibinfo{author}{Z.~Mao}, \bibinfo{author}{B.~Wang}, \bibinfo{author}{L.~Guo},
\newblock \bibinfo{title}{Knowledge graph embedding: A survey of approaches and applications},
\newblock \bibinfo{journal}{IEEE Transactions on Knowledge and Data Engineering} \bibinfo{volume}{29} (\bibinfo{year}{2017}) \bibinfo{pages}{2724--2743}.
\bibitem[{Kulmanov et~al.(2021)Kulmanov, Smaili, Gao, and Hoehndorf}]{kulmanov2021semantic}
\bibinfo{author}{M.~Kulmanov}, \bibinfo{author}{F.~Z. Smaili}, \bibinfo{author}{X.~Gao}, \bibinfo{author}{R.~Hoehndorf},
\newblock \bibinfo{title}{Semantic similarity and machine learning with ontologies},
\newblock \bibinfo{journal}{Briefings in Bioinformatics} \bibinfo{volume}{22} (\bibinfo{year}{2021}) \bibinfo{pages}{bbaa199}.
\bibitem[{Li et~al.(2022)Li, Ding, Zhi, Gu, Wang et~al.}]{li2022identification}
\bibinfo{author}{J.~Li}, \bibinfo{author}{J.~Ding}, \bibinfo{author}{D.~Zhi}, \bibinfo{author}{K.~Gu}, \bibinfo{author}{H.~Wang}, et~al.,
\newblock \bibinfo{title}{Identification of type 2 diabetes based on a ten-gene biomarker prediction model constructed using a support vector machine algorithm},
\newblock \bibinfo{journal}{BioMed Research International} \bibinfo{volume}{2022} (\bibinfo{year}{2022}).
\bibitem[{Mansoori et~al.(2018)Mansoori, Ghaedi, Sadatamini, Vahabpour, Rahimipour, Shanaki, Saeidi, and Kazerouni}]{mansoori2018downregulation}
\bibinfo{author}{Z.~Mansoori}, \bibinfo{author}{H.~Ghaedi}, \bibinfo{author}{M.~Sadatamini}, \bibinfo{author}{R.~Vahabpour}, \bibinfo{author}{A.~Rahimipour}, \bibinfo{author}{M.~Shanaki}, \bibinfo{author}{L.~Saeidi}, \bibinfo{author}{F.~Kazerouni},
\newblock \bibinfo{title}{Downregulation of long non-coding rnas linc00523 and linc00994 in type 2 diabetes in an iranian cohort},
\newblock \bibinfo{journal}{Molecular biology reports} \bibinfo{volume}{45} (\bibinfo{year}{2018}) \bibinfo{pages}{1227--1233}.
\bibitem[{Kazerouni et~al.(2020)Kazerouni, Bayani, Asadi, Saeidi, Parvizi, and Mansoori}]{kazerouni2020type2}
\bibinfo{author}{F.~Kazerouni}, \bibinfo{author}{A.~Bayani}, \bibinfo{author}{F.~Asadi}, \bibinfo{author}{L.~Saeidi}, \bibinfo{author}{N.~Parvizi}, \bibinfo{author}{Z.~Mansoori},
\newblock \bibinfo{title}{Type2 diabetes mellitus prediction using data mining algorithms based on the long-noncoding rnas expression: a comparison of four data mining approaches},
\newblock \bibinfo{journal}{BMC bioinformatics} \bibinfo{volume}{21} (\bibinfo{year}{2020}) \bibinfo{pages}{1--13}.
\bibitem[{Saeidi et~al.(2018)Saeidi, Ghaedi, Sadatamini, Vahabpour, Rahimipour, Shanaki, Mansoori, and Kazerouni}]{saeidi2018long}
\bibinfo{author}{L.~Saeidi}, \bibinfo{author}{H.~Ghaedi}, \bibinfo{author}{M.~Sadatamini}, \bibinfo{author}{R.~Vahabpour}, \bibinfo{author}{A.~Rahimipour}, \bibinfo{author}{M.~Shanaki}, \bibinfo{author}{Z.~Mansoori}, \bibinfo{author}{F.~Kazerouni},
\newblock \bibinfo{title}{Long non-coding rna ly86-as1 and hcg27\_201 expression in type 2 diabetes mellitus},
\newblock \bibinfo{journal}{Molecular biology reports} \bibinfo{volume}{45} (\bibinfo{year}{2018}) \bibinfo{pages}{2601--2608}.
\bibitem[{Zhu et~al.(2020)Zhu, Zhu, Liu, Jiang, Chen, Cheng, Cheng et~al.}]{zhu2020gene}
\bibinfo{author}{H.~Zhu}, \bibinfo{author}{X.~Zhu}, \bibinfo{author}{Y.~Liu}, \bibinfo{author}{F.~Jiang}, \bibinfo{author}{M.~Chen}, \bibinfo{author}{L.~Cheng}, \bibinfo{author}{X.~Cheng}, et~al.,
\newblock \bibinfo{title}{Gene expression profiling of type 2 diabetes mellitus by bioinformatics analysis},
\newblock \bibinfo{journal}{Computational and Mathematical Methods in Medicine} \bibinfo{volume}{2020} (\bibinfo{year}{2020}).
\bibitem[{Chang et~al.(2020)Chang, Bala{\v{z}}evi{\'c}, Allen, Chawla, Brandt, and Taylor}]{chang2020benchmark}
\bibinfo{author}{D.~Chang}, \bibinfo{author}{I.~Bala{\v{z}}evi{\'c}}, \bibinfo{author}{C.~Allen}, \bibinfo{author}{D.~Chawla}, \bibinfo{author}{C.~Brandt}, \bibinfo{author}{R.~A. Taylor},
\newblock \bibinfo{title}{Benchmark and best practices for biomedical knowledge graph embeddings},
\newblock in: \bibinfo{booktitle}{Proceedings of the conference. Association for Computational Linguistics. Meeting}, volume \bibinfo{volume}{2020}, \bibinfo{organization}{NIH Public Access}, \bibinfo{year}{2020}, p. \bibinfo{pages}{167}.
\bibitem[{Bordes et~al.(2013)Bordes, Usunier, Garcia-Dur\'{a}n, Weston, and Yakhnenko}]{bordes2013translating}
\bibinfo{author}{A.~Bordes}, \bibinfo{author}{N.~Usunier}, \bibinfo{author}{A.~Garcia-Dur\'{a}n}, \bibinfo{author}{J.~Weston}, \bibinfo{author}{O.~Yakhnenko},
\newblock \bibinfo{title}{Translating embeddings for modeling multi-relational data},
\newblock in: \bibinfo{booktitle}{Proceedings of NIPS 2013}, \bibinfo{publisher}{Curran Associates Inc.}, \bibinfo{address}{Red Hook, NY, USA}, \bibinfo{year}{2013}, p. \bibinfo{pages}{2787–2795}.
\bibitem[{Yang et~al.(2015)Yang, tau Yih, He, Gao, and Deng}]{yang2014embedding}
\bibinfo{author}{B.~Yang}, \bibinfo{author}{W.~tau Yih}, \bibinfo{author}{X.~He}, \bibinfo{author}{J.~Gao}, \bibinfo{author}{L.~Deng}, \bibinfo{title}{Embedding entities and relations for learning and inference in knowledge bases}, \bibinfo{year}{2015}.
\bibitem[{Ristoski and Paulheim(2016)}]{ristoski2016rdf2vec}
\bibinfo{author}{P.~Ristoski}, \bibinfo{author}{H.~Paulheim},
\newblock \bibinfo{title}{{{RDF2Vec}: {RDF} graph embeddings for data mining}},
\newblock in: \bibinfo{booktitle}{Proceedings of the 15th International Semantic Web Conference}, \bibinfo{publisher}{Springer International Publishing}, \bibinfo{address}{Cham, Switzerland}, \bibinfo{year}{2016}, pp. \bibinfo{pages}{498--514}.
\bibitem[{Clough et~al.(2024)Clough, Barrett, Wilhite, Ledoux, Evangelista, Kim, Tomashevsky, Marshall, Phillippy, Sherman et~al.}]{clough2024ncbi}
\bibinfo{author}{E.~Clough}, \bibinfo{author}{T.~Barrett}, \bibinfo{author}{S.~E. Wilhite}, \bibinfo{author}{P.~Ledoux}, \bibinfo{author}{C.~Evangelista}, \bibinfo{author}{I.~F. Kim}, \bibinfo{author}{M.~Tomashevsky}, \bibinfo{author}{K.~A. Marshall}, \bibinfo{author}{K.~H. Phillippy}, \bibinfo{author}{P.~M. Sherman}, et~al.,
\newblock \bibinfo{title}{Ncbi geo: archive for gene expression and epigenomics data sets: 23-year update},
\newblock \bibinfo{journal}{Nucleic Acids Research} \bibinfo{volume}{52} (\bibinfo{year}{2024}) \bibinfo{pages}{D138--D144}.
\bibitem[{Preisner and Paulheim(2022)}]{preisner2022universal}
\bibinfo{author}{P.~Preisner}, \bibinfo{author}{H.~Paulheim},
\newblock \bibinfo{title}{Universal preprocessing operators for embedding knowledge graphs with literals}  (\bibinfo{year}{2022}).
\bibitem[{Consortium(2021)}]{GO2021}
\bibinfo{author}{G.~Consortium},
\newblock \bibinfo{title}{{The Gene Ontology resource: enriching a {GO}ld mine}},
\newblock \bibinfo{journal}{Nucleic Acids Research} \bibinfo{volume}{49} (\bibinfo{year}{2021}) \bibinfo{pages}{D325--D334}.
\bibitem[{Huntley et~al.(2015)Huntley, Sawford, Mutowo-Meullenet, Shypitsyna, Bonilla, Martin, and O'Donovan}]{GOA2015}
\bibinfo{author}{R.~P. Huntley}, \bibinfo{author}{T.~Sawford}, \bibinfo{author}{P.~Mutowo-Meullenet}, \bibinfo{author}{A.~Shypitsyna}, \bibinfo{author}{C.~Bonilla}, \bibinfo{author}{M.~J. Martin}, \bibinfo{author}{C.~O'Donovan},
\newblock \bibinfo{title}{{The {GOA} database: gene ontology annotation updates for 2015}},
\newblock \bibinfo{journal}{Nucleic Acids Research} \bibinfo{volume}{43} (\bibinfo{year}{2015}) \bibinfo{pages}{D1057--D1063}.
\bibitem[{Szklarczyk et~al.(2021)Szklarczyk, Gable, Nastou, Lyon, Kirsch, Pyysalo, Doncheva, Legeay et~al.}]{string2021}
\bibinfo{author}{D.~Szklarczyk}, \bibinfo{author}{A.~L. Gable}, \bibinfo{author}{K.~C. Nastou}, \bibinfo{author}{D.~Lyon}, \bibinfo{author}{R.~Kirsch}, \bibinfo{author}{S.~Pyysalo}, \bibinfo{author}{N.~T. Doncheva}, \bibinfo{author}{M.~Legeay}, et~al.,
\newblock \bibinfo{title}{{The {STRING} database in 2021: customizable protein--protein networks, and functional characterization of user-uploaded gene/measurement sets}},
\newblock \bibinfo{journal}{Nucleic Acids Research} \bibinfo{volume}{49} (\bibinfo{year}{2021}) \bibinfo{pages}{D605--D612}.
\bibitem[{Quinlan(1986)}]{quinlan1986induction}
\bibinfo{author}{J.~R. Quinlan},
\newblock \bibinfo{title}{Induction of decision trees},
\newblock \bibinfo{journal}{Machine learning} \bibinfo{volume}{1} (\bibinfo{year}{1986}) \bibinfo{pages}{81--106}.
\bibitem[{GSE(2014)}]{GSE30208}
\bibinfo{title}{{Series GSE30208}},
\newblock \bibinfo{year}{2014}. \URLprefix \url{https://www.ncbi.nlm.nih.gov/geo/query/acc.cgi?acc=GSE30208}.
\bibitem[{Kallionp{\"a}{\"a} et~al.(2014)Kallionp{\"a}{\"a}, Elo, Laajala, Mykk{\"a}nen, Ricano-Ponce, Vaarma, Laajala, Hy{\"o}ty, Ilonen, Veijola et~al.}]{kallionpaa2014innate}
\bibinfo{author}{H.~Kallionp{\"a}{\"a}}, \bibinfo{author}{L.~L. Elo}, \bibinfo{author}{E.~Laajala}, \bibinfo{author}{J.~Mykk{\"a}nen}, \bibinfo{author}{I.~Ricano-Ponce}, \bibinfo{author}{M.~Vaarma}, \bibinfo{author}{T.~D. Laajala}, \bibinfo{author}{H.~Hy{\"o}ty}, \bibinfo{author}{J.~Ilonen}, \bibinfo{author}{R.~Veijola}, et~al.,
\newblock \bibinfo{title}{Innate immune activity is detected prior to seroconversion in children with hla-conferred type 1 diabetes susceptibility},
\newblock \bibinfo{journal}{Diabetes} \bibinfo{volume}{63} (\bibinfo{year}{2014}) \bibinfo{pages}{2402--2414}.
\bibitem[{GSE(2012)}]{GSE15932}
\bibinfo{title}{{Series GSE15932}},
\newblock \bibinfo{year}{2012}. \URLprefix \url{https://www.ncbi.nlm.nih.gov/geo/query/acc.cgi?acc=GSE15932}.
\bibitem[{GSE(2014)}]{GSE55098}
\bibinfo{title}{{Series GSE55098}},
\newblock \bibinfo{year}{2014}. \URLprefix \url{https://www.ncbi.nlm.nih.gov/geo/query/acc.cgi?acc=GSE55098}.
\bibitem[{Yang et~al.(2015)Yang, Ye, Wang, Gao, Liu, Hong, Wang, Gu, and Ning}]{yang2015decreased}
\bibinfo{author}{M.~Yang}, \bibinfo{author}{L.~Ye}, \bibinfo{author}{B.~Wang}, \bibinfo{author}{J.~Gao}, \bibinfo{author}{R.~Liu}, \bibinfo{author}{J.~Hong}, \bibinfo{author}{W.~Wang}, \bibinfo{author}{W.~Gu}, \bibinfo{author}{G.~Ning},
\newblock \bibinfo{title}{Decreased mi r-146 expression in peripheral blood mononuclear cells is correlated with ongoing islet autoimmunity in type 1 diabetes patients 1},
\newblock \bibinfo{journal}{Journal of diabetes} \bibinfo{volume}{7} (\bibinfo{year}{2015}) \bibinfo{pages}{158--165}.

\end{thebibliography}

\end{document}